# TURNOVER PREDICTION OF SHARES USING DATA MINING TECHNIQUES: A CASE STUDY

Shashaank D.S1, Sruthi.V2, Vijayalashimi M.L.S3 and Shomona Garcia Jacob4

[1]Department of Computer Science and Engineering, SSNCE, Chennai, India.

shashaank.sivakumar@gmail.com

[2]Department of Computer Science and Engineering, SSNCE, Chennai, India.

sruthivenkatesh1@gmail.com

[3]Department of Computer Science and Engineering, SSNCE, Chennai, India.

vijayalakshimisethuraman@gmail.com

[4]Department of Computer Science and Engineering, SSNCE, Chennai, India.

shomonagj@ssn.edu.in

**ABSTRACT**

*Predicting the Total turnover of a company in the ever fluctuating Stock market has always proved to be a precarious situation and most certainly a difficult task at hand. Data mining is a well-known sphere of Computer Science that aims at extracting meaningful information from large databases. However, despite the existence of many algorithms for the purpose of predicting future trends, their efficiency is questionable as their predictions suffer from a high error rate. The objective of this paper is to investigate various existing classification algorithms to predict the turnover of different companies based on the Stock price. The authorized dataset for predicting the turnover was taken from www.bsc.com and included the stock market values of various companies over the past 10 years. The algorithms were investigated using the 'R' tool. The feature selection algorithm, Boruta, was run on this dataset to extract the important and influential features for classification. With these extracted features, the Total Turnover of the company was predicted using various algorithms like Random Forest, Decision Tree, SVM and Multinomial Regression. This prediction mechanism was implemented to predict the turnover of a company on an everyday basis and hence could help navigate through dubious stock markets trades. An accuracy rate of 95% was achieved by the above prediction process. Moreover, the importance of the stock market attributes was established as well.*

.

## 1. INTRODUCTION

Prediction of stock market prices, its rise and fall of values has constantly proved to be a perilous task mainly due to the volatile nature of the market[1-3]. However data mining techniques and other computational intelligence techniques have been applied to achieve the same over the years. Some of the approaches undertaken included the use of decision tree algorithm, concepts of neural networks and Midas[4-6]. However through this paper, a comparative study was conducted to estimate and predict the turnover of companies that include Infosys, Sintex, HDFC and Apollo hospitals using various machine learning algorithms such as Random Forest, Decision Tree, Support Vector Machine and Multinomial Logistic Regression. In order to estimate the performance of the aforementioned machine learning algorithms in predicting the turnover, a confusion matrix was also constructed with respect to the dataset. Based on the predictions made by each of the algorithms with respect to the total turnover for a company (on an everyday basis), an accuracy rate was estimated for each of them from the number of true positives/negatives and false positives/negatives. A brief review of the state-of-the-art in predicting stock market share data is given below.

## 2. RELATED WORK

The objective of any nation at large is to enhance the lifestyle of common man and that is the driving force to undertake research to predict the market trends [7-9]. In the recent decade, much research has been done on neural networks to predict the stock market changes [10].

Matsui and Sato [12] proposed a new evaluation method to dissolve the over fitting problem in the Genetic Algorithm (GA) training. On comparing the conventional and the neighbourhood evaluation they found the new evaluation method to be better than the conventional one in terms of performance. Gupta, Aditya, and Dhingra [13] proposed a stock market prediction technique based on Hidden Markov Models. In that approach, the authors considered the fractional change in stock value and the intra-day high and low values of the stock to train the continuous Hidden Markov Model (HMM). Then this HMM is used to make a Maximum a Posteriori decision over all the possible stock values for the next day. The authors applied this approach on several stocks, and compared the performance to the existing methods. Lin, Guo, and Hu[14] proposed a SVM based stock market prediction system .This system selected a good feature subset, evaluated stock indicator and controlled over fitting on stock market tendency prediction. The authors tested this approach on Taiwan stock market datasets and found that the proposed system surpassed the conventional stock market prediction system in terms of performance.

## 3. PROPOSED STOCK TUROVER PREDICTION FRAMEWORK

The stock turnover prediction framework proposed in this paper is portrayed in Figure 1. The basic methodology involved Data Collection, Pre-processing, Feature Selection and Classification, each of which is explained below.

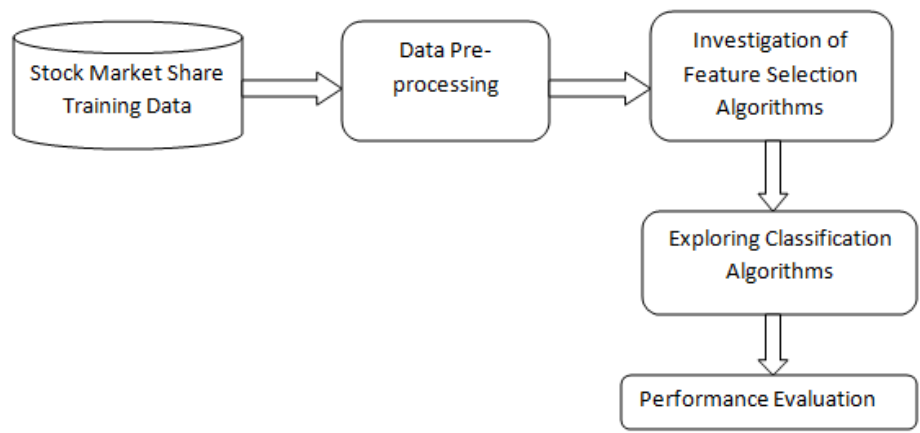

Figure 1: Stock Turnover Prediction framework

The dataset utilised for predicting the turnover was taken from www.bsc.com which included the stock market values of companies including Infosys, HDFC, Apollo Hospitals and Sintex, over the past 10 years.

## 3.1 Data Processing

Initially, all the records with missing values were removed from the dataset in order to improve the accuracy of the prediction. Then the data was further partitioned into two parts:

**Training data (d.t)**: It is the data with which the machine is trained. Various classification algorithms are trained on this data. 60% of the data is taken as training data.

**Validation data (d.v)**: It is the data which is used for the purpose of cross-validation. It is used to find the accuracy rate of each algorithm. The remaining 40% of the data is taken as validation data.

In order to apply the classification algorithms, the data was first sorted according to the turnover. Then the total turnover was discretised into:

A - 58,320 to 18,291,986

B – 18,296,597 to 37,731,606

C – 37,749,751 to 121,233,543

D – 121,245,870 to 300,360,881

E- 300,465,316 to 19,085,311,470

Also, the company features were converted into dummy variables (0's/1's) to help the prediction process easier.

The stock market data was characterised by attributes described in Table 1. The stock market starts at 9:15 in the morning and ends at 3:30 in the afternoon. The attributes described in Table 1 are recorded within this time frame.

Table 1. Stock Market Share Data – Attribute Description

| S.NO | ATTRIBUTE | DESCRIPTION |
|---|---|---|
| 1. | Open price | The first traded price during the day or in the morning. |
| 2. | High price | The highest traded price during the day. |
| 3. | Low price | The lowest price traded during the day. |
| 4. | Close price | The last price traded during the day. |
| 5. | WAP | Weighted average price during the day. |
| 6. | No of shares | The total number of shares done during the day. |
| 7. | No of trades | No of trades is the total no of transactions during the day. |
| 8. | Deliverable quantity. | The quantity that can be delivered at the end of the day. |
| 9. | Spread high low | Range of High price and low prices. |
| 10. | Spread close open | Range of close and low prices. |
| 11. | Company | The name of the company that handles the shares. |
| 12. | Total turn over | Turnover is the total no of shares traded X Price of each share sold. |
| 13. | Date | The date for which the above attributes are |

|   |   | recorded. |
|---|---|---|

Once the data was pre-processed, the important features to make an accurate prediction were identified by the process of feature selection.

### 3.2 Feature Selection

In order to estimate the possible influence of each of the above attributes on the predicted turnover, Boruta algorithm in R tool [15] was used. Boruta is a machine learning algorithm used to find relevant and important features for a given dataset i.e used to solve the minimal-optimal problem. The minimal – optimal problem is an often found situation today where most of the variables in a dataset are irrelevant to its classification. This problem gives rise to various disadvantages including over consumption of resources, slowdown of machine learning algorithms and most importantly, decrease in accuracy yielded by the same. Additionally, Boruta is a wrapper algorithm built around the Random Forest algorithm( implemented in the R package RandomForest) such that in every iteration the algorithm removes the irrelevant or less important features or attributes on the basis of the results rendered by a series of statistical tests.

The Boruta algorithm follows the following steps:

- The information system is expanded by adding duplicates of all attributes. These duplicates are known as shadow attributes.
- The added attributes are shuffled and the randomForest algorithm is run on the expanded information system. The resultant Z scores are noted.
- The Z score of every attribute is considered and the maximum Z score among all the shadow attributes (MZSA) is estimated. Further a hit value is assigned to every attribute that possesses a Z score better than MZSA.
- For each shadow attribute with undetermined importance perform a two-sided test of equality with the MZSA is conducted.
- All the attributes which have significantly lower importance than MZSA are considered to be 'unimportant' and permanently removed them from the information system.
- Similarly those attributes that having higher importance when considered alongside MZSA are considered to be important.
- All duplicates from the information system are removed.
- This procedure is repeated until the level of importance is assigned for all the attributes.

*3.2 Sample code:*

*f.la <- Total_Turnover ~.*

*at.select <- Boruta (formula = f.la, data = d.t)*

*at.select$finalDecision*

*And, the graph obtained is as given in figure 2*

### 3.3 Classification

Classification [16-17] is the process of finding a set of models that describe and distinguish data classes. This is done to achieve the goal of being able to use the model to predict the class whose label is unknown. The classification phase involved the execution of the classification algorithms to identify the best performing algorithm. The classification accuracy obtained by percentage split as discussed in the data pre-processing phase, was calculated and a comparison was drawn among the classifiers. The algorithms that yielded the highest accuracy is described below.

*Random Forest*

In random forest [18-19] a randomly selected set of attributes is used to split each node. Every node is split using the best split among a subset of predictors that are deliberately chosen randomly at the node. This is in contrast to the methodology followed in standard tress in which each node is split using the best split among all attributes available in the dataset. Further new values are predicted by aggregating and collating the predictions of the various decision trees constructed.

Random forest represents an ensemble model / algorithm as it derives its final prediction from multiple individual models. These individual models could be of similar or different type. However, in the case of Random Forest, the individual models are of the same type – decision trees.

*Sample code:*

```
f.la <- Total_Turnover ~.
dt.fit <- randomForest(formula = f.la,data = d.t)
dt.fit.v <- predict (object = dt.fit,newdata = d.v ,type = 'class')
```

The Random Forest algorithm yielded 95.08% accuracy with all the 12 features, the results of which are discussed in the ensuing section.

The comparative performances of the feature selection and classification algorithms are discussed below.

## 4. RESULT ANALYSIS

The results analysis is discussed in two sections. The former section elaborates on the feature selection process while the latter section makes a detailed analysis on the performance of the classification algorithms.

### 4.1   Performance Analysis of Feature Selection

As discussed before, the Boruta package was utilized for performing feature selection on the pre-processed training data. The graphical representation of the importance of the features and their role in enhancing the classification accuracy is portrayed in Figure 2.

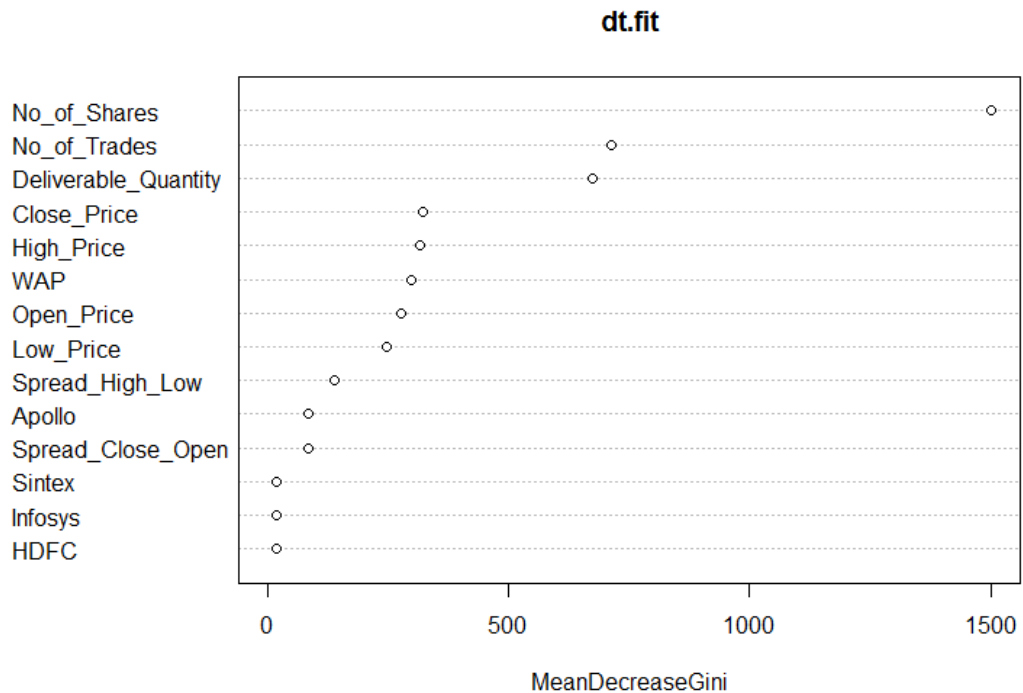

**Figure 2: Attribute Importance in Share Turnover Prediction – Boruta Package**

Once the important features were identified, the next phase involved predicting the turnover from the features in order to estimate the probable combination of attributes that yield a high turnover.

### 4.2 Performance Analysis of Classification Algorithms

Each of the classification algorithms were first trained using the training data which contained 60% of the dataset. Then the remaining 40% was used for the purpose of cross-validation. From the prediction produced by each of the algorithms, the confusion matrix was obtained. Further the accuracy rate was determined by the formula:

$$\text{Accuracy rate} = \frac{\text{No. of correctly classified observations}}{\text{Total No. of observations}} \times 100$$

**Table 2. Comparative Performance of Classification Algorithms**

| S.No | Classification Algorithms | Accuracy (%) |
|---|---|---|
| 1 | Random Forest | 95.08 |
| 2 | Decision tree – PARTY | 89.5 |
| 3 | Decision tree- Rpart | 82.3 |
| 4 | SVM | 75.9 |
| 5 | MLR | 73.55 |

The graphical representation of the total turnover prediction of the companies is given in Figure 3.

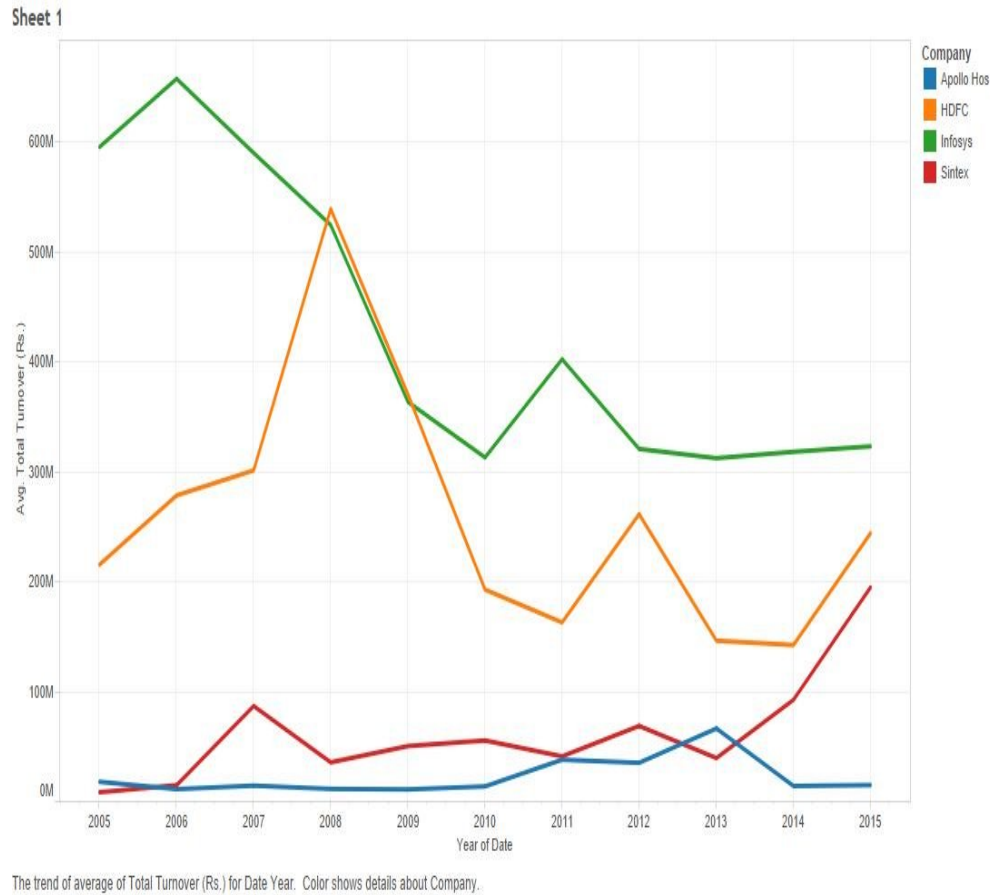

**Figure 3 .Overall Prediction of Turnover for the Companies**

The line graph was plotted between the Average turnover for each year and the corresponding year itself. The details of the graph shown in Figure 3 are as follows:

- Apollo hospital: Initially increased from in 18,020,386 in 2005 to 66,527,438 in 2013 and then decreased to 14,995,685 in 2015.
- Infosys: Initially decreased from 595,911,899 in 2005 to 313,287,090 in 2010, increased to 402,652,597 in 2011 and further decreased to 323,405,496 in 2015.
- Sintex: Increased with a few variations from 8,495,478 in 2005 to 194,974,362 in 2015.
- HDFC: Initially increased from 251,751,341 in 2005 to 539,536,979 in 2008 and then decreased to 244,309,549 in 2015.

The bar graph was plotted between 9831 opening prices grouped by total turnover with the discretized value of the total turnover. The graph clearly stated that there total turnover did not increase linearly with respect to the sum of the No of shares. For example, the No of shares for D was 529,957,607 whereas, for E it is 617,959,679.

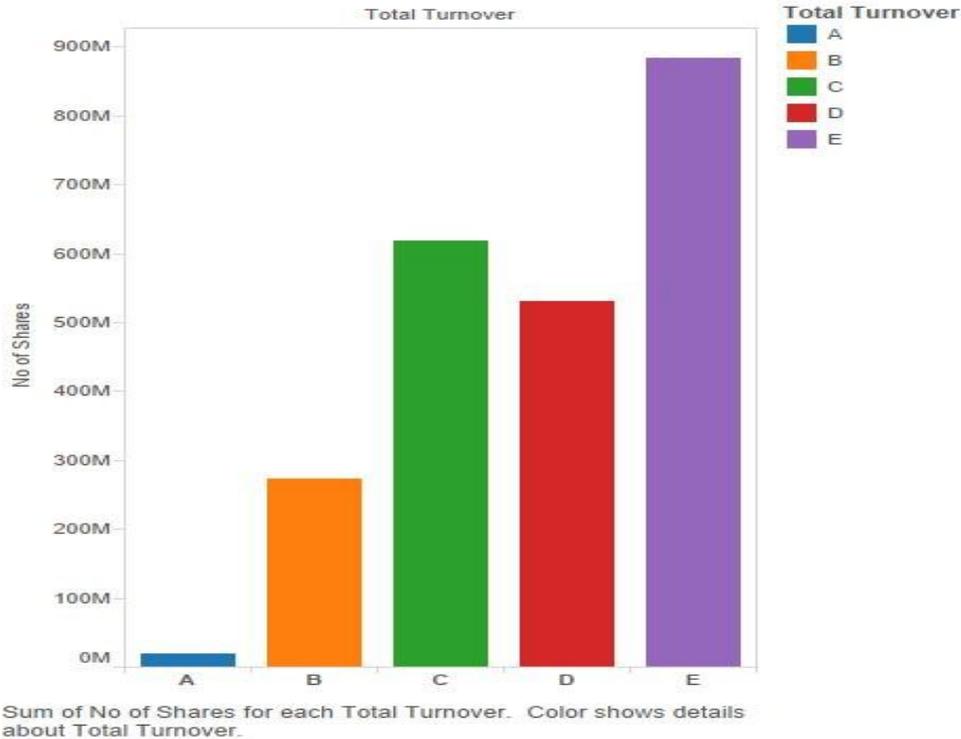

**Figure 4. Variation of Total_Turnover with Respect to No_Of_Shares**

It is evident from the result analysis that almost all the features are required to accurately predict the total turnover of a company. Moreover the Random Forest algorithm has proved to accurately predict the turnover for the real-time share data of different companies which gives the lead to investigate many other boosting and ensemble techniques to enhance the prediction accuracy.

## 5. CONCLUSION

Application of data mining techniques to predict turnover based on stock market share data is an emerging area of research and is expected to be instrumental in moulding the country's economy by predicting possible investment trends to increase turnover. In view of this, an efficient way of implementing the Random Forest algorithm is proposed in order to mitigate the risks involved in predicting the turnover of a company. It was also identified that all features involved in the stock marker share data were essential for prediction. An accuracy rate of 95% was achieved in the prediction process. This accuracy rate was much higher than those obtained before. Hence we believe that further research using computational methodologies to predict turnover on a daily basis based on share market data will reveal better and more interesting patters for investments.

## Authors

Shashaank D.S is currently pursuing B.E computer Science and Engineering in SSN College of Engineering Chennai, India. He is doing research in the field of machine learning and is interested in speech processing.

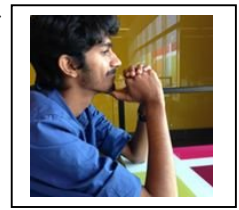

Sruthi.V is currently pursuing B.E computer Science and Engineering in SSN College of Engineering Chennai, India. She is doing research in the field of machine learning.

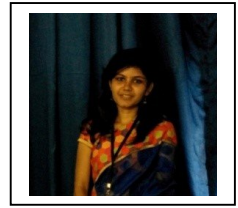

Vijayalakshimi M.L.S is currently pursuing B.E computer Science and Engineering in SSN College of Engineering Chennai, India. She is doing research in the field of machine learning.

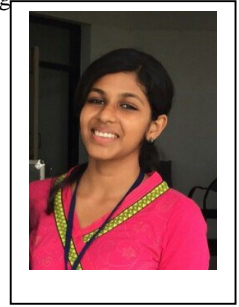

Dr. Shomona Gracia Jacob is Associate Professor, Department of CSE, SSN College of Engineering, Chennai, India. She completed Ph.D at Anna University in the area of Biological and Clinical Data Mining. She has more than 30 publications in International Conferences and Journals to her credit. Her areas of interest include Data Mining, Bioinformatics, Machine Learning, and Artificial Intelligence. She has reviewed many research articles on invitation from highly reputed refereed journals. She is currently guiding under-graduate and post-graduate projects in the field of data mining and intelligent systems.

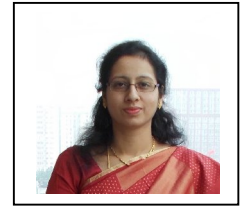